\documentclass{article}
\usepackage{ijcai22}

\usepackage{times}

\usepackage{soul}
\usepackage{url}
\usepackage[hidelinks]{hyperref}
\usepackage[utf8]{inputenc}
\usepackage[small]{caption}
\usepackage{graphicx}
\usepackage{amsmath}
\usepackage{booktabs}
\usepackage{xcolor}
\usepackage{verbatim}
\usepackage{enumitem}

\usepackage[draft,inline,nomargin,index]{fixme}

\fxsetup{theme=color,mode=multiuser}
\FXRegisterAuthor{mp}{mmp}{\color{red}Maitreya}

\usepackage{latexsym}

\title{Reasoning about Actions over Visual and Linguistic Modalities: A Survey}

\author{
Shailaja Keyur Sampat$^1$\footnote{Corresponding Author. This survey will be periodically updated with latest works in this area.}\and Maitreya Patel$^1$ \and Subhasish Das$^1$ \and Yezhou Yang$^1$ \And Chitta Baral$^1$%\and
\affiliations $^1$Arizona State University\\
\emails
\{ssampa17, mpatel57, sdas48, yz.yang, chitta\}@asu.edu
}

\begin{document}

\maketitle

\begin{abstract}
`Actions' play a vital role in how humans interact with the world and enable them to achieve desired goals. As a result, most common sense (CS) knowledge for humans revolves around actions. While `Reasoning about Actions \& Change' (RAC) has been widely studied in the Knowledge Representation community, it has recently piqued the interest of NLP and computer vision researchers. This paper surveys existing tasks, benchmark datasets, various techniques and models, and their respective performance concerning advancements in RAC in the vision and language domain. Towards the end, we summarize our key takeaways, discuss the present challenges facing this research area, and outline potential directions for future research.  

%'Actions' play a vital role in our day-to-day life as they enable us to interact with the world. As a result, a significant amount of human's reasoning and common knowledge revolve around actions. 
%it is relatively less explored in NLP and Vision communities. 
\end{abstract}

% lessons learned from the topic can contribute to new ideas and visions that can stimulate the research community to pursue new directions, e.g., new problems.
% clear why the topic is important and what its intended or existing applications are.
\section{Overview}

Humans interact with their environment to accomplish desired goals. Object manipulation (i.e., performing `actions' over the objects) is a fundamental concept that makes this interaction possible. On the other hand, we often encounter dynamic states of the world. In such situations, we anticipate underlying actions that are responsible for those changes. Thus, actions in their simplest form have the power to change the state of the world. As a result, a significant amount of commonsense (CS) knowledge we use in our day-to-day life revolves around actions. For example, we can easily perceive when a stack of dishes or a Jenga tower will topple; and we can estimate that to what extent the grocery bag should be filled so that it does not tear or crush the contents \cite{battaglia2013simulation}.

%For example: If we want to eat a banana (goal), we have to peel the banana (action). If we fry a potato (action), the potato will become golden and crisp (resulting state). If we observe the ice cubes (resulting state), we know that it was obtained by freezing (action) the water (previous state).

Reasoning about actions is important for humans as it helps us to predict if a sequence of actions will lead us to achieve the desired goal; to explain observations in terms of what actions may have taken place; and to diagnose faults i.e. identifying actions that may have resulted in undesirable situation \cite{baral2010reasoning}. As we are developing autonomous agents that can assist us in performing everyday tasks, they would also require to interact with complex environments and make rational decisions. As pointed out by \cite{davis2015commonsense}, imagine a guest asks a robot for a glass of wine; if the robot sees that the glass is broken or has a dead cockroach inside, it should not pour the wine and serve it. Similarly, if a cat runs in front of a house-cleaning robot, the robot should neither run it over nor sweep it up nor put it away on a shelf. Hence, the ability of artificial agents to perform reasoning and integrate CS knowledge about actions is highly desirable. 

As mentioned above, rapid physical inferences are central to how humans interact with the world, and it is important to understand its computational aspects as we make progress in Artificial Intelligence (AI). As a result, Reasoning about Action and Change (RAC) has been a long-established research problem, since the rise of AI. \cite{mccarthy1960programs} were the first to emphasize the importance of reasoning about the effect of the actions. They developed an advice taker system that can do deductive reasoning about scenarios such as ``going to the airport from home" requires ``walking to the car" and ``driving the car to the airport". %In the followup works, \cite{mccarthy1963situations} developed  Situation Calculus as a tool to reason about actions and proposed the well-known `Frame Problem' \cite{McCHay69}.

In over four decades of research, the knowledge representation and reasoning (KR\&R) community has been successful in developing promising solutions to RAC problems. One of our co-authors co-presented an IJCAI 2019 tutorial \cite{AISur} on this topic. In addition to discussions on historical developments in RAC, they provided a concise summary of various kinds of commonsense knowledge centered around actions required to solve the Winograd Schema Challenge \cite{levesque2012winograd}. That being a stepping-stone, here, we broaden our horizon, specifically with the recent works in NLP and Computer Vision related to RAC. %Specifically, that hold five assumptions- single-agent performing actions, perfect knowledge of the world, and actions being atomic, deterministic, only cause for changes in the world \cite{reiter2001knowledge}.  Extensions and overcoming limitations of the above aspects %i.e. continuous domain, domain with multiple agents, imperfect knowledge of the world, possibility of probabilistic or simultaneous events are actively being explored. 
\begin{table*}[]
\centering
\small
\begin{tabular}{@{}ll@{}}
\toprule
\multicolumn{1}{c}{\textbf{Reasoning Type (with corresponding Inputs and Outputs)}} & \multicolumn{1}{c}{\textbf{Example}} \\ \midrule
\begin{tabular}[c]{@{}l@{}}\textit{\textbf{Temporal Prediction}}\\ \quad Initial State, Action(s) $\rightarrow$ Resulting State\end{tabular} & \begin{tabular}[c]{@{}l@{}}Potato peeled and cut into pieces, Frying\\ $\rightarrow$ Potato pieces are crisp and golden\end{tabular} \\ \midrule
\begin{tabular}[c]{@{}l@{}}\textbf{\textit{Temporal Explanation}}\\ \quad Initial State$^*$, Resulting State $\rightarrow$ Explanation about Action(s)\\ \\ \quad Action(s), Resulting State $\rightarrow$ Explanation about Initial State\end{tabular} & \begin{tabular}[c]{@{}l@{}}\\Carrot with the skin removed $\rightarrow$ Peeling action \\ performed on the whole carrot causes skin removal \\ Freezing, Ice $\rightarrow$ Water if freezed turns into ice\end{tabular} \\ \midrule
\begin{tabular}[c]{@{}l@{}}\textit{\textbf{Goal driven Action Prediction/Planning}}\\ \quad Initial State$^*$, Resulting State $\rightarrow$ Action(s)\end{tabular} & \begin{tabular}[c]{@{}l@{}}Cake batter in the baking pan, The cake is baked\\ $\rightarrow$ Put the pan into oven for 20-25 minutes at 350F\end{tabular} \\ \midrule
\begin{tabular}[c]{@{}l@{}}\textbf{\textit{Temporal Dependency}}\\ \quad  Action(s) $\rightarrow$ Action(s) performed Before/After\end{tabular} & \begin{tabular}[c]{@{}l@{}}Changing of a tire $\rightarrow$ Flattening of the tire (after)\\ Eating a banana $\rightarrow$ Peeling the banana (before)\end{tabular} \\ \bottomrule
\end{tabular}
\caption{\centering Some key types of reasoning about actions in vision and language with definitions and examples (optional inputs are indicated by $^*$).}
\label{tab:restype}
\vspace{-3mm}
\end{table*}

With the success of deep learning based methods across a wide range of vision and language tasks, NLP and vision communities have shown growing interest in RAC. \cite{weston2015aicomplete} was one of the earliest works to incorporate reasoning about actions. They set up a suite of 20 toy tasks (bAbI) which aims at solving several types of reasoning provided as natural language contexts. Though their key focus was not on actions, many tasks required understanding of verbs (such as go, pick, drop, move and their synonyms) performed by different subjects in order to solve the task. The work \cite{forbes2017verb} developed a system that can take unstructured natural language text and generate relative physical knowledge (of size, weight, speed, strength and rigidness) and implications of actions. For example, given ``x barged into y", one can conclude that ``x is smaller than y".
%with the promising performance of fine-grained object detectors and large-scale language models,

The core interest of vision and language researchers lies in determining whether neural models would be able to reason about actions, provided a dataset with linguistic and/or visual inputs. For example, ATOMIC \cite{sap2019atomic} is a knowledge base of neural network generated linguistic commonsense inferences about actions. Provided a text snippet, they generate three kinds of inferences around it; actions that could have happened before, actions that can happen after, and subject's intents at present. Similarly, in the vision domain, \cite{Gokhale2019CVPRWorkshops} attempted AI planning task using neural networks over a pair of images in blocks world setting.

In gist, an ability to reason about actions is crucial for AI agents in closing the gap between human-level performance and machine performance in NLP, vision, and robotics tasks. In this paper, we aim to survey advances in this research direction with a focus on visual and linguistic modalities\footnote{Scope of this work includes assessing the capability of vision \& language architectures to do reasoning about actions and we do not make comparisons with models that incorporate intuitive physics.}. The remainder of the paper is structured four-fold; First, we provide a detailed taxonomy of various reasoning types related to actions in the vision \& NLP domains. Second, we discuss benchmark datasets and evaluation mechanisms designed to systematically assess a machine's capability to perform such reasoning. We then discuss various techniques/models that have been developed to tackle this challenge and their performance.  
Finally, we present the challenges facing this research area and outline potential directions for future research. 

%\cite{shanahan1997solving}
%\cite{DBLP:journals/corr/WestonBCM15}

%\section{Background and Scope}

% existing work should be structured and should contain a discussion (aiming for instance at comparing alternative approaches, or identifying open questions and challenges).
\vspace{-0.2cm}
\section{Taxonomy of Reasoning Types}
In this section, we describe five main types of reasoning associated with actions based on our observations from the vision \& NLP literature. Table \ref{tab:restype}  provides a concise summary of each type with corresponding inputs/outputs and relevant examples. Some clarifications regarding Table 1 are as follows; First, the reasoning types described above are defined based on what aspect of action understanding is of key focus and can be different from the downstream task used for systematic evaluation. Second, a state can consist of one more object in any possible modality (in our case, states can be expressed as visuals- images or videos, structured representations, or linguistic constructs of any length) and actions may be atomic or composite. This assumption holds throughout this paper. Third, oftentimes, initial state and actions are combined into a single modality, specifically when expressed in language. 

Actions are omnipresent and an integral part of our lives. Over the time, with our own experience and observing other people, we accumulate a vast amount of commonsense knowledge related to actions. We observe six most frequent types of commonsense knowledge i.e. preconditions associated with actions, co-occurrence of objects helpful for performing actions, location at which specific actions are performed, useful tools and materials, attributes associated with objects after performing actions, and intents behind actions. We provide examples of each knowledge category in Table \ref{tab:comsen}. %In section \ref{sec:kres}, we discuss some existing resources that can be useful in obtaining commonsense knowledge related to actions that can be incorporated into models.

\begin{table}[h]
%\textit{\textbf{Common Knowledge about Actions}}\\
\centering
\small
\begin{tabular}{@{}ll@{}}
\toprule \multicolumn{1}{c}{\textbf{Knowledge Type}} & \multicolumn{1}{c}{\textbf{Example}} \\ \midrule
\begin{tabular}[c]{@{}l@{}}Preconditions\\ Co-occurrence\\ Locations \\ Tools/Materials \\ Object attributes \\ Intents \end{tabular} & \begin{tabular}[c]{@{}l@{}}\textbf{Liquid} objects can undergo \textit{pouring}\\ \textbf{Knife} is often mentioned with \textit{cut} \\ \textit{Brushing}  takes place in \textbf{washroom}\\ \textbf{Oven} is useful for \textit{baking} \\ \textit{Painting} an object changes its \textbf{color} \\ One \textit{drinks} water if they are \textbf{thirsty}\end{tabular} \\ \bottomrule
\end{tabular}
\caption{Examples of Knowledge that is helpful in performing reasoning about actions (\textbf{boldface} represents the knowledge for a given type, \textit{italic} represents action to which  knowledge is associated}
\label{tab:comsen}
\vspace{-5mm}
\end{table}

\begin{comment}

Note: $<>$- predicted, () - given \\
Note:  \\
\begin{itemize}
    \item Action preceding the given action\\
$<$action1$>$, (action2): action1 happens before given action2

\item Action following the given action\\
(action1), $<$action2$>$: action2 happens after given action1

\item Predict effect/outcome of performing actions \\
(actions), $<$state$>$: predict state that will be achieved if we perform given actions \\
(state1), (actions), $<$state2$>$: predict state2 that will be achieved if we perform given actions on the state1

\item
Predict actions that will lead to a given state/outcome \\
$<$actions$>$, (state): predict actions that can be performed to achieve desired state \\
(state1), $<$actions$>$, (state2): predict actions if performed on state1, we end up in state2 \\

Note: Many QA datasets will be part of this (QA is performed over predicted state essentially) 

\item Commonsense related to actions  \\

\end{itemize}
\end{comment}

\section{Benchmark Tasks and Datasets}
\label{sec:datasets}

In this section, we describe various tasks requiring reasoning about actions and existing datasets that are useful in assessing the capability of models to perform those tasks. %For better organization, we divide the tasks/datasets into- vision only, text only,  vision+language, and other/applications. %In the end, we discuss some large-scale resources knowledge sources related to actions.

\vspace{-0.1cm}
\subsection{Vision-Only Tasks}

Here, we discuss handful of studied tasks that involve visuals (images/videos) and reasoning about actions based on them\footnote{Though action recognition may seem a relevant task here, we forbear to include them here as they do not require any further ``reasoning" and due to space constraints}. %There are only a few works considering the fact that our focus is on reasoning about actions beyond what can be simply perceived.  

\begin{figure}[ht!]
    {\centering
    (a) \includegraphics[width=0.8\linewidth]{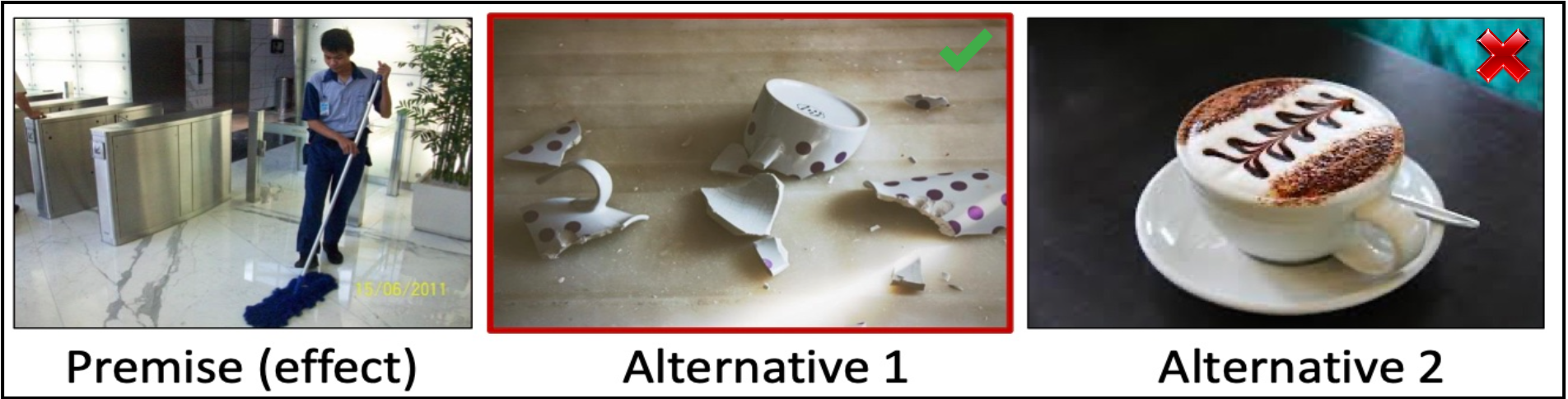}\\}
    {\centering
    (b) \includegraphics[width=0.8\linewidth]{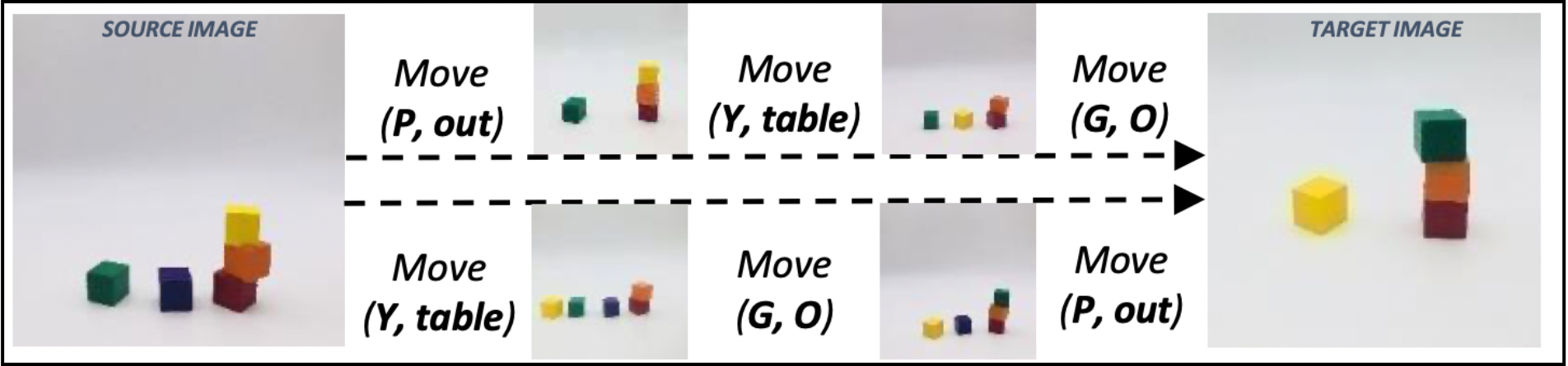}\\ }
    {\centering
    (c) \includegraphics[width=0.8\linewidth]{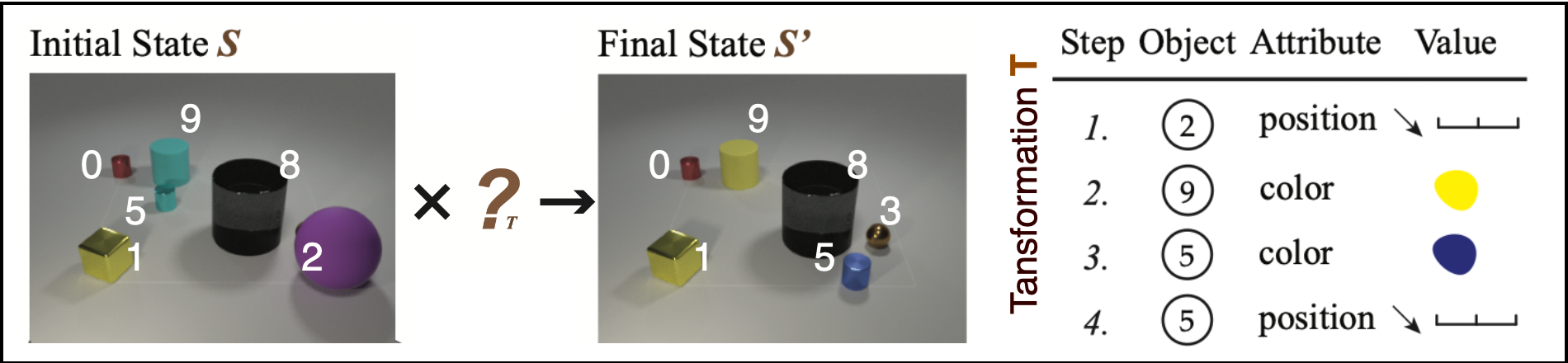}\\}

    \caption{Vision-only tasks requiring reasoning about actions; (a) Causality understanding task of \protect\cite{yeo2018visual}; (b) and (c) Image based AI-planning tasks of \protect\cite{Gokhale2019CVPRWorkshops} and  \protect\cite{hong2021transformation} respectively}
    \label{fig:visex}
    \vspace{-3mm}
\end{figure}

\paragraph{Causality and Actions}
Though the research on causality is a well-established direction in both vision and NLP, it is highly related to understanding action as actions play an important role as a cause for many of the events. VisualCOPA \cite{yeo2018visual} was proposed as a visual task to encourage the development of models that can understand cause-effect relations. Specifically, given a premise image and two alternative images, the task is to identify the more plausible alternative with their commonsense causal context. A significant portion of VisualCOPA requires understanding actions depicted in the images and then reasoning about actions that might have taken place in the past, actions that might take place in the future, and commonsense knowledge such as ``people prefer drinking white wine with the seafood" \cite{yeo2018visual}. An example is shown in Figure \ref{fig:visex}(b). 

\paragraph{Vision based AI-planning}
Recently, there have been attempts to tackle AI-planning task using neural networks. Specifically, provided a start image and a goal image, the task is to generate a sequence of actions or a plan that if applied on the start image can lead to object configurations in the goal image.  \cite{hong2021transformation} proposed TRANCE dataset where transformations between two images are captured in terms of changes in color, shape, size, material, or position of objects as a sequence of (object, attribute, value) tuples. BIRD dataset proposed by \cite{Gokhale2019CVPRWorkshops}  is very similar from the task formulation point of view but has important distinctions. First, in the BIRD dataset only blocks with five colors are present whereas the TRANCE dataset has more variety in terms of object attributes. Second, in the BIRD dataset object stacking is possible (as per Figure  \ref{fig:visex}(a)), which is not the case with TRANCE. As a result, there are stricter conditions associated with performing actions e.g. object can be moved only if there is no other object on its top.

\vspace{-0.1cm}
\subsection{Language-Only Tasks}

\begin{figure}[ht!]
    {\centering
    (a) \includegraphics[width=0.8\linewidth]{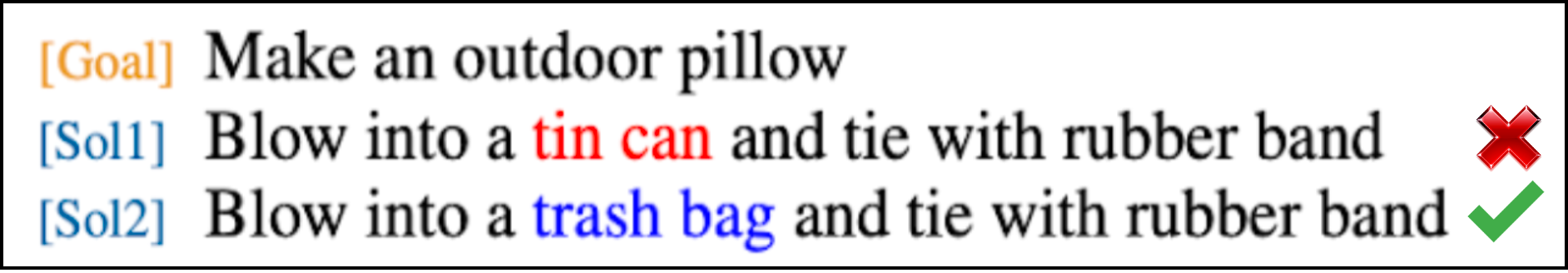}\\}
    {\centering
    (b) \includegraphics[width=0.8\linewidth]{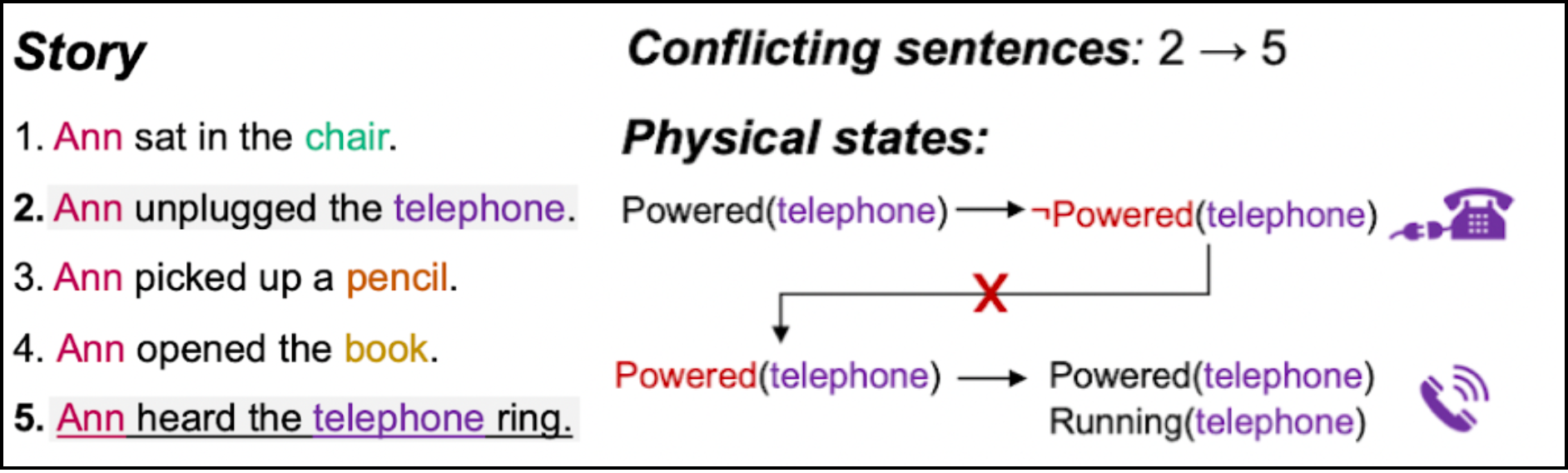}\\}
     {\centering
    (c) \includegraphics[width=0.8\linewidth]{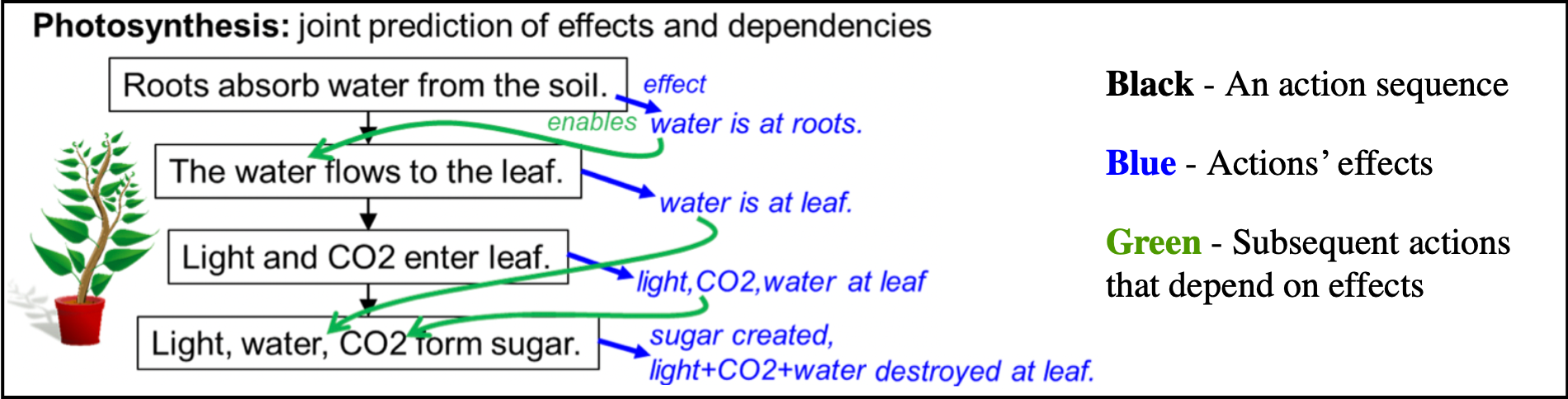}\\}
     {\centering
    (d) \includegraphics[width=0.8\linewidth]{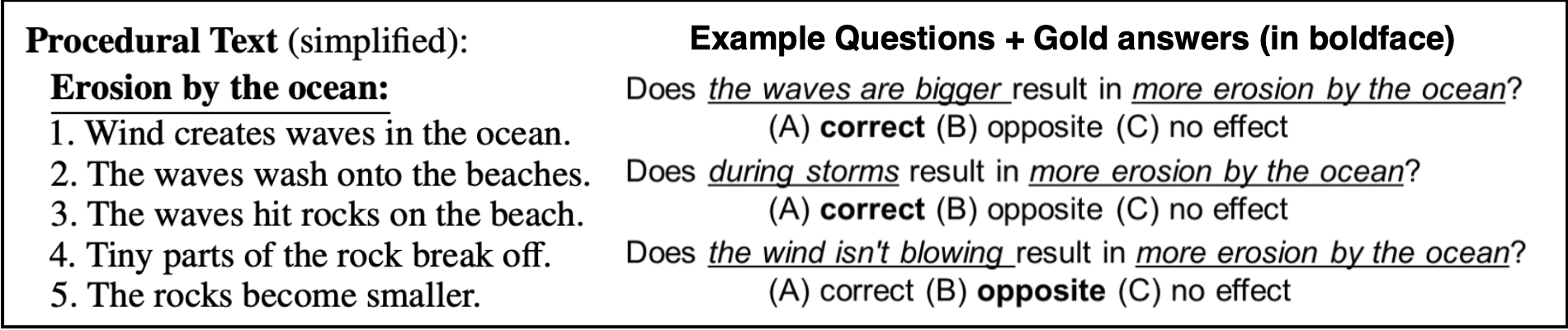}\\}
    {\centering
    (e) \includegraphics[width=0.8\linewidth]{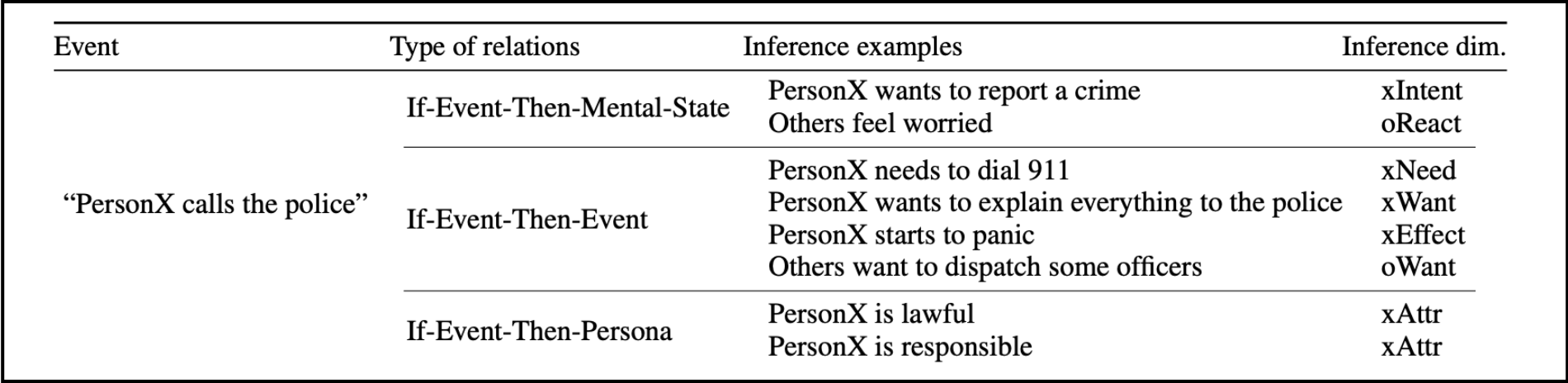}\\}
    \caption{Language-only tasks requiring reasoning about actions; (a) Goal-driven physical commonsense task \protect\cite{bisk2020piqa};  (b) Conflict detection task of \protect\cite{storks2021tiered} requiring physical commonsense reasoning; (c) Action explanation in procedural texts \protect\cite{dalvi2019everything}; (d) What-if reasoning about actions increase/decrease \protect\cite{tandon2019wiqa}; (e) Commonsense inference about events \protect\cite{sap2019atomic}}
    \label{fig:langex}
    \vspace{-3mm}
\end{figure}

In this section, we discuss tasks that involve language-based reasoning about actions. Pure language-based tasks are suitable when high-level goal descriptions are to be mapped with a set of actions or for explainability purposes i.e. to provide justification about the choice of action or commonsense knowledge that is useful to make conclusions. Figure \ref{fig:langex} shows some representative examples of such tasks.  As states become more complicated and involve many different objects, it becomes hard to convey every single detail (about object's position, attributes such as colors, sizes, textures, etc.) through text and requires complex descriptions to refer to objects to avoid possible ambiguity. Following are existing text-only datasets that focus on action reasoning;

\paragraph{High-level reasoning about Actions}
 The TRIP \cite{storks2021tiered} is a dataset for physical commonsense reasoning (specifically verb causality, preconditions, and rules of intuitive physics) expressed in natural language. One task is about predicting precondition (solidity- of potato as for being cut) and effect (in pieces- resulting state of potato after cutting) for a given sentence ``John cut the cooked potato in half" and the entity ``potato". Whereas, another task is about determining conflicting pairs of sentences among a story that demonstrates implausible physical commonsense (Ex. Ann unplugged the telephone, Ann heard the telephone ring). PIQA \cite{bisk2020piqa} is a benchmark for goal-driven prediction of the actions (along with the choice of correct objects and necessary tools) formulated as a two-way task. An interesting aspect of this dataset is that many examples demonstrate unconventional usage of the objects ex. a cotton swab can be used to apply an eyeshadow in absence of the brush. 
 
 \paragraph{Explainability and Commonsense about Actions}
ATOMIC \cite{sap2019atomic} is a language-based commonsense knowledge base for what-if reasoning. Given a text snippet, it can fetch relevant commonsense about events that could have happened before, events that can happen after, and the subject's intents at present. In \cite{dalvi2019everything}, the authors extended their procedural text understanding dataset ProPara by incorporating a new task of explaining the effects of actions and their role in performing subsequent actions. WIQA \cite{tandon2019wiqa} is a testbed for what-if reasoning over natural language contexts. Provided a procedural paragraph, the task is to answer questions of the kind ``Does changeX result in changeY?" (where X and Y are two events from the paragraph) with 3-way answer choices- correct, opposite, or no effect.    
 
\vspace{-0.1cm}
\subsection{Vision+Language Tasks}

\paragraph{Change Captioning}
%bisk-etal-2016-natural,
The work of \cite{Bisk2018LearningIS} proposed a task where given a pair of images in a 3D blocks world setting, the model learns to predict spatial actions in natural language such that if executed, it will convert one image into another. Later \cite{jhamtani2018learning} proposed the Spot-the-Difference task which aims at generating a multi-sentence description of differences between two images. To overcome their limitation about not being robust against distractors and no viewpoint shifts considered between images, \cite{park2019viewpoint} proposed CLEVR-Change dataset. The underlying task is to generate a caption describing the difference between a pair of synthetic images. 

\paragraph{Commonsense Retrieval/Generation}
\cite{yang2018commonsense} proposed the ActionExplanation task; first given an image, the model has to identify action being performed as a classification task and second, rank commonsense triplets according to their relevance with the image. In VisualComet, \cite{park2020visualcomet} attempted to generate commonsense inferences such as events that could have happened before/after, and subject's intents at present in a given image. 

\paragraph{Instruction Following} 
Language-guided image manipulation is an emerging research direction in vision+language. While a majority of datasets involve object and attribute level scene manipulations, CSS \cite{vo2019composing} and CRIR \cite{chen2020graph} have explored action level changes. Another relevant task under this category is vision-and-language navigation (VLN) \cite{mattersim,chen2018touchdown,nguyen2019vnla}, where an agent navigates in a visual environment to find a goal location by following linguistic instructions. All the above datasets (namely R2R, TouchDown and AskNav) include visuals, natural language instructions, and a set of actions that can be performed to achieve desired goals.
Further, ALFRED \cite{shridhar2020alfred} increased the complexity level of the VLN task for agents by adding long, compositional tasks. The task comprises of dealing with longer action sequences, complex action space, and language, that are closely related to real-world situations. 
\begin{figure}[ht!]
    {\centering
    (a)
    \includegraphics[width=0.8\linewidth]{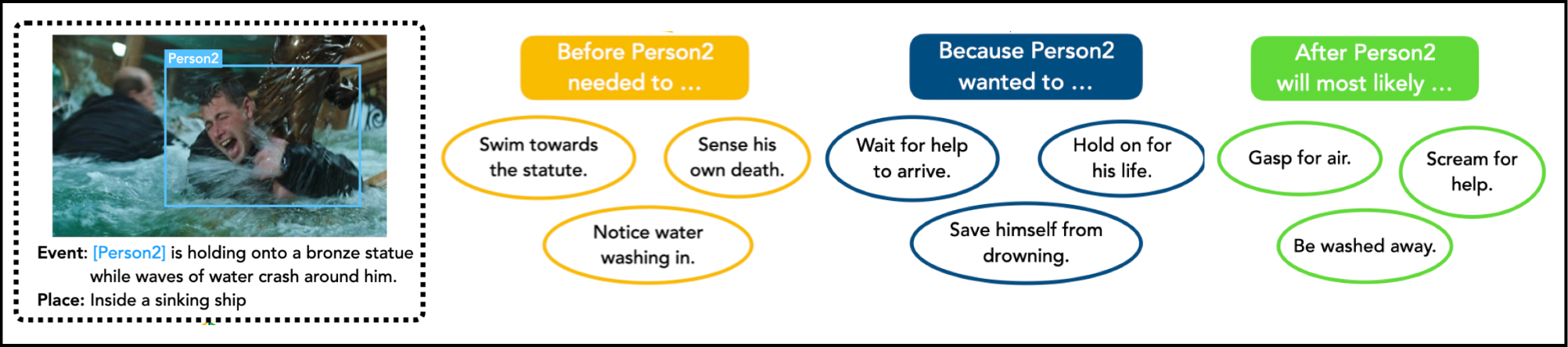} \\}
    {\centering
    (b) \includegraphics[width=0.8\linewidth]{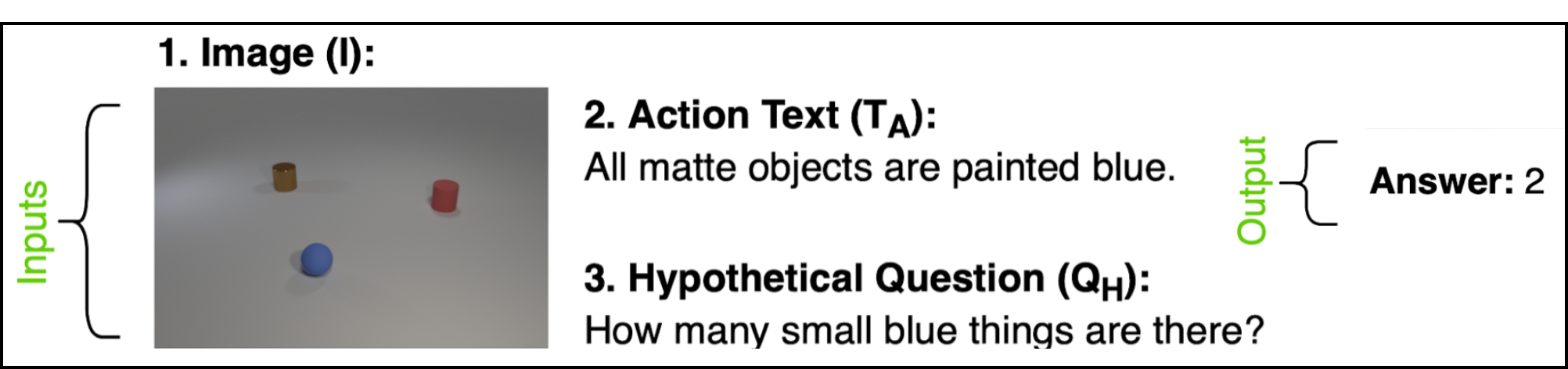}\\}
     {\centering
    (c) 
    \includegraphics[width=0.8\linewidth]{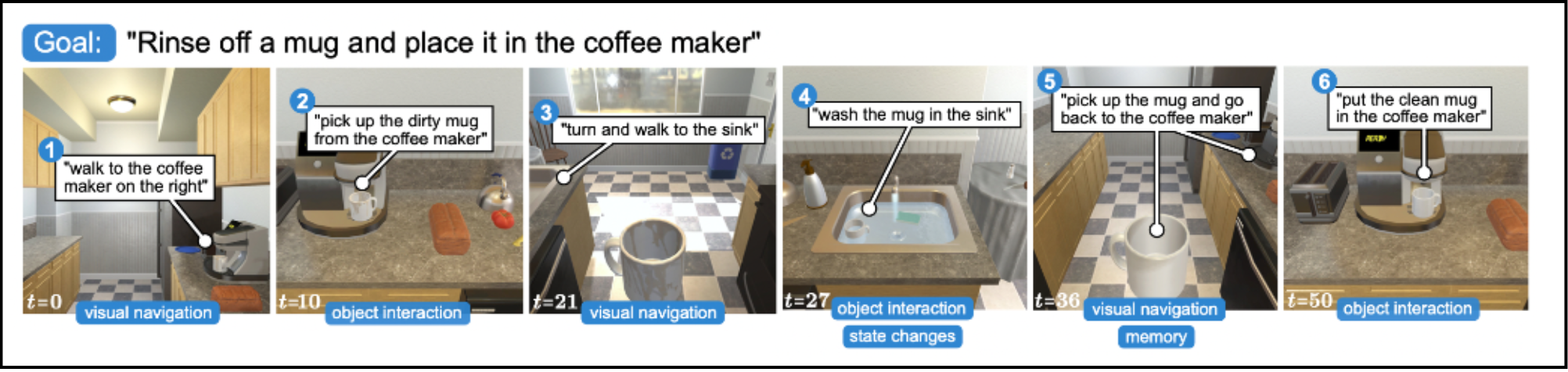} \\}
    {\centering
    (d) 
    \includegraphics[width=0.8\linewidth]{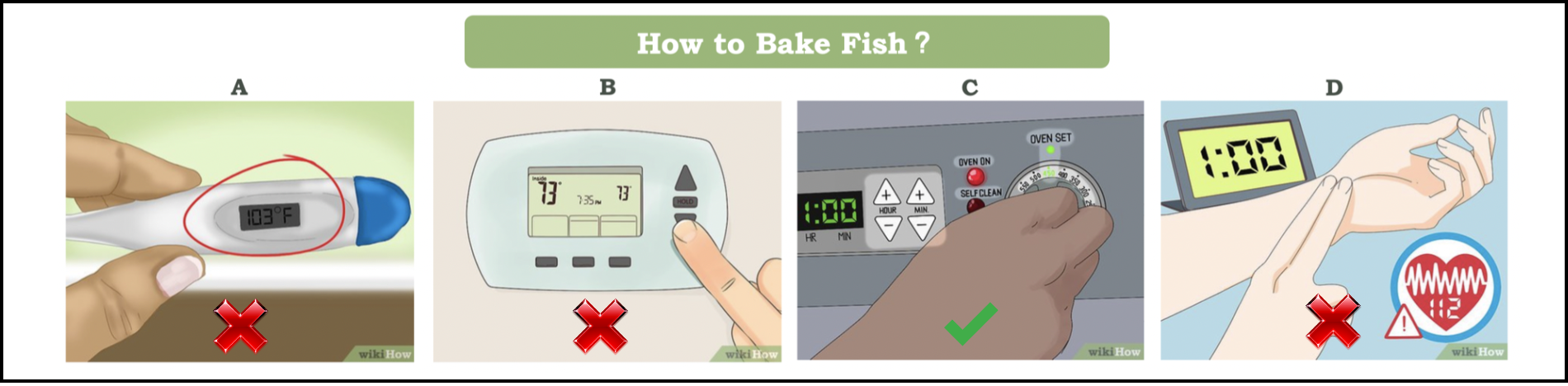} \\}
    {\centering
    (e) 
    \includegraphics[width=0.8\linewidth]{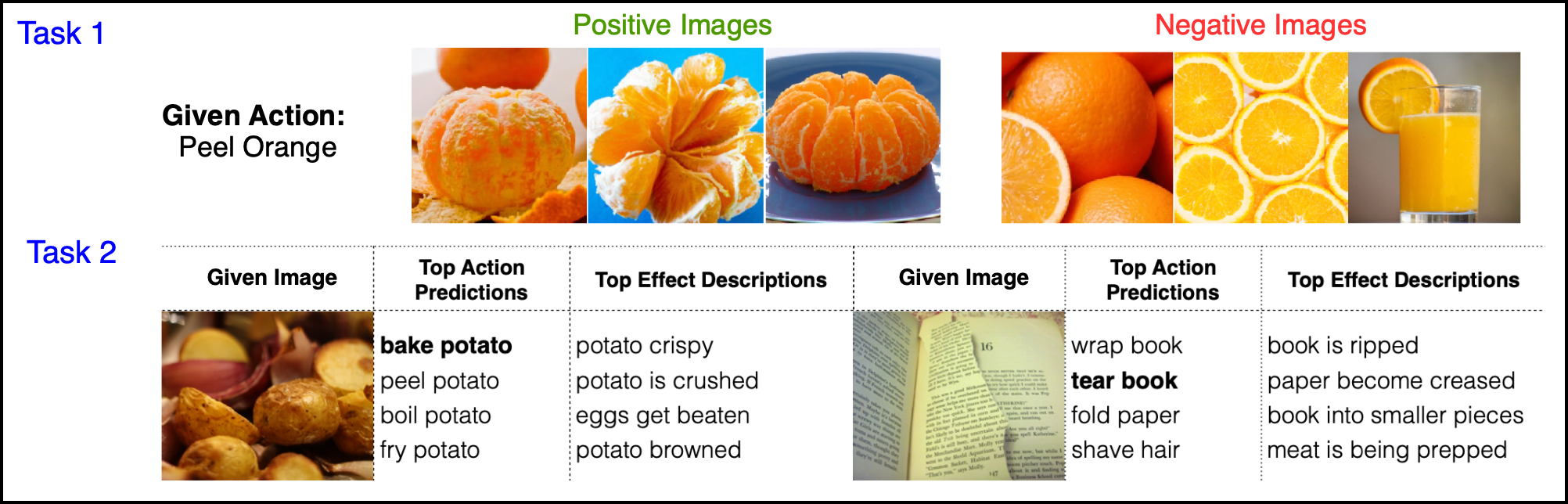} \\}
    \caption{Vision-language tasks requiring reasoning about actions; (a) Commonsense inference generation about events before, after and intents at present from an image \protect\cite{park2020visualcomet}; (b) VQA about hypothetical actions \protect\cite{sampat2021clevr_hyp}; (c) VLN task of   \protect\cite{shridhar2020alfred}; (d) Goal-driven action identification \protect\cite{yang2021visual}; (e) Action conditioned image-retrieval,  image conditioned text-retrieval focusing on actions-effects  \protect\cite{gao2018action}} 
    \label{fig:vislangex}
    \vspace{-3mm}
\end{figure}

\paragraph{Visual Question Answering}
In TIWIQ dataset \cite{wagner2018answering}, provided rendered table-top scenes, the task is to generate a response to what-if question when an action (push, rotate, remove or drop) is performed on an object. 
Recently, \cite{sampat2021clevr_hyp} created CLEVR\_HYP dataset. It shares similarities with TIWIQ for having rendered images, limited action types, and question-answering tasks. However, the key difference is that in CLEVR\_HYP, the impact of actions on the overall scene is required and not just object-level changes. Though vision-language tasks of RecipeQA \cite{yagcioglu2018recipeqa} do not explicitly involve actions, it is necessary to have an understanding of actions to figure out an incoherent image for a recipe or arrange images in a meaningful order to prepare a recipe. CLEVRER \cite{yi2019clevrer} incorporates understanding videos where object collisions take place, followed by the QA task of four kinds- descriptive (e.g., ``what color"), explanatory (``what is responsible for"), predictive (``what will happen next"), and counterfactual (``what if"). A considerable subset of VCR \cite{zellers2019recognition} dataset involves answering questions about actions specifically, including hypothetical, mental and temporal aspects.

%% testing image resolution and readability
% references need to be added for all papers
%\cite{sampat2021clevr_hyp} \cite{bisk2020piqa} \cite{girdhar2019cater} \cite{shridhar2020alfred} \cite{lohmann2020learning}

\begin{comment}

\begin{figure*}[h]
    \centering
    \includegraphics[width=0.85\textwidth]{images/v3.png}
    \caption{Overview of existing datasets that require reasoning about actions}
    \label{fig:my_label}
\end{figure*}

\end{comment}

%\paragraph{Goal-driven Action Prediction}

\paragraph{Miscellaneous}
\cite{yang2021visual} proposed the Visual Goal-Step Inference (VGSI) task and created a dataset using wikiHow as a resource. Given a high-level textual goal and a set of candidate images, the model learns to identify the ones which constitute a reasonable step towards the given goal.  
\cite{gao2018action} collected a dataset for learning action-effects in a multi-modal setting. One task focuses on predicting action (from a set of candidates) that is a likely cause for a state shown in the image. Whereas, the other task is about predicting the effect of a given verb-noun pair by identifying suitable images from a pool. 
\cite{isola2015discovering} collected a dataset of objects, scenes, and materials, each being in a variety of transformed states. Their proposed tasks are three-fold; One, given a noun, discovering relevant transformations it can undergo (e.g. a tomato can undergo slicing, cooking, etc.). Second, given an image, assign states that can be perceived (e.g., sliced, raw, etc.). Third, arranging a set of images with a common object according to their attributes between a pair of antonymic states (e.g. images of bacon from raw to cooked, tomatoes from unripe to ripe, etc.). 

\subsection{Other Tasks}
Learning a mapping from natural language instructions to a sequence of actions to be performed in a visual environment is a common task in robotics \cite{kanu2020following}. In another recent work, \cite{chen2020graph} proposed a task of scene-graph modification conditioned on a natural language command. They crowdsourced annotations (supporting three types of graph operations- insert, delete or substitute) over scene-graphs. \cite{gaddy2019pre} proposed the task of learning a mapping from natural language instructions to state transitions while being data efficient. Given an initial state (as a structured representation) and natural language instruction to perform some action over it, the task is to predict the resulting state (as a structured representation). 

%\paragraph{Robotics, Game playing AI- } 
%{\textcolor{green}{need to add more works here!}}

\begin{comment}

\begin{itemize}
    \item Classical AI Tasks \\
    Planning etc.
    \item Vision and Language
    \item Applications \\
    Robotics, Game playing AI etc. 
\end{itemize}

\end{comment}

\subsection{Knowledge Resources}
\label{sec:kres}
 We discuss a few large-scale resources which can be helpful in creating datasets related to action reasoning or can be used as a supplementary knowledge base while developing models to solve tasks described above.

 %\paragraph{Activity-related Datasets} with detailed annotations \\
 %{\textcolor{green}{need to add more works here!}}

 \paragraph{Instructables.com, Wikihow} are websites with a crowd-sourced collection of instructions for a wide range of tasks in day-to-day life. Each instance contains visual information (images or videos), descriptions of each step, and a list of required tools. There are two important benefits of using the above sources to create datasets; While most tools and materials are expected to be familiar with most people, a significant portion of instances demonstrate atypical or unusual uses of everyday objects. For example, a hair net can be tied on the end of the vacuum to find small objects lost in a room. Second, textual instructions in these resources build on one another which allows one to capture insights about underlying assumptions, preconditions that need to be met to begin the task, and postconditions that define the success of the task.
 
 \paragraph{ConceptNet} \cite{speer2017conceptnet} is a large-scale knowledge graph about concepts that humans use frequently and relationships among them, expressed in various natural languages. In the latest version of ConceptNet, there are 34 possible relations among concepts. A subset of those relationships (specifically PartOf, HasA, UsedFor, CapableOf, Causes, HasSubevent, HasPrerequisite, HasProperty, MotivatedByGoal, ObstructedBy, CausesDesire, MadeOf, ReceivesAction) can be useful in deriving various kind of knowledge related to actions as a chain of reasoning. For example, provided ConceptNet relations- (cake CreatedBy bake) and (oven CapableOf bake), we can derive conclusions such as `one requires oven if they want to bake a cake'. 
 
\begin{table}[!t]
\centering
\resizebox{\columnwidth}{!}{
\begin{tabular}{@{}lll@{}}
\toprule
\multicolumn{1}{c}{\textbf{\begin{tabular}[c]{@{}c@{}}Task/\\ Dataset\end{tabular}}} & \multicolumn{1}{c}{\textbf{\begin{tabular}[c]{@{}c@{}}SOTA Model \\   Performance$^\#$ \\\end{tabular}}} & \multicolumn{1}{c}{\textbf{\begin{tabular}[c]{@{}c@{}}Human \\ Performance\end{tabular}}} \\ \midrule
VisualCOPA & 56.1\% & - \\ \midrule
BIRD & 88.5\% & 100\% \\ \midrule
TRANCE & 70.0\% & 89.1\% \\ \midrule \midrule
TRIP & 75.2\% & - \\ \midrule
PIQA & 90.1\% & 94.9\% \\ \midrule
% ATOMIC & {\color[HTML]{FF0000} } &  \\ \midrule
ProPara Expl. & 60.7\% & 80.7\% \\ \midrule
WIQA & 78.5\% & 96.3\% \\ \midrule \midrule
% 3D Blocks World &  &  \\ \midrule
Spot-the-Difference & \begin{tabular}[c]{@{}l@{}}0.09 Bleu-4, 0.29 Rouge-L\end{tabular} & \begin{tabular}[c]{@{}l@{}}0.08 Bleu-4, 0.31 Rouge-L\end{tabular} \\ \midrule
CLEVR-Change & \begin{tabular}[c]{@{}l@{}}0.62 Bleu-4, 0.50 METEOR \\ 1.16 CIDEr, 0.53 Rouge-L \end{tabular} & - \\ \midrule
% Action Expl. &  &  \\ \midrule
VisualComet & 38.7\% Acc@50 & - \\ \midrule
CSS & 99.8\% Recall@1 & - \\ \midrule
CRIR & 98.4\% Recall@1 & - \\ \midrule
R2R & \begin{tabular}[c]{@{}l@{}}78\% Success\\ 686.54 Length (m)\end{tabular} & \begin{tabular}[c]{@{}l@{}}86\% Success\\ 11.85 Length (m)\end{tabular} \\ \midrule
Touchdown & \begin{tabular}[c]{@{}l@{}}19.35 TC ($\uparrow$)\\ 16.87 SPD ($\downarrow$)\\ 18.77 SED ($\uparrow$)\end{tabular} & 92\% SDR@80px \\ \midrule
% AskNav &  &  \\ \midrule
ALFRED & 34.0\% unseen & 91\% unseen \\ \midrule
TIWIQ & 64.0\% & 95.3\% \\ \midrule
CLEVR\_HYP & 70.5\% & 98.4\% \\ \midrule
RecipeQA & \begin{tabular}[c]{@{}l@{}}83.6\% VCloze\\ 70.3\% VOrdering\\ 80.1\% VCoherence\end{tabular} & \begin{tabular}[c]{@{}l@{}}77.6\% VCloze\\ 64\% VOrdering\\ 81.6\% VCoherence\end{tabular} \\ \midrule
CLEVRER & 95.2\% & - \\ \midrule
VGSI & 72.8\% & 84.5\% \\ \midrule
ActionEffect & \begin{tabular}[c]{@{}l@{}}75.0\% Top5 VRetrieval\\ 84.3\% Top5 TRetrieval\end{tabular}  & - \\
 \bottomrule
\end{tabular}}
\caption{State-of-the-art (SOTA) model performance and Human upperbound on vision-language datasets that require reasoning about Actions (T-Textual, V-Visual, \#-To the best of our knowledge)}
\label{table:sota}
\vspace{-3mm}
\end{table}

\section{Evaluation}

In this section, we discuss evaluation mechanisms designed to systematically assess a machine's capability to perform action-effect reasoning. 

   \paragraph{Textual/Visual Classification} 
     measures are frequently used when the task has pre-defined categories which can be associated with the inputs or the most relevant text/visual has to be selected from plausible choices. accuracy, Precision, Recall, F-1 Score, and Word-similarity are various methods to evaluate a classification task. A significant number of temporal prediction and temporal explanation reasoning tasks (e.g. multiple-choice tasks) use classification measures. 
     
  \paragraph{Textual/Visual Retrieval} style measure is suitable for downstream tasks where given a query, the task is to identify texts/visuals from a given pool according to their relevance with the query. Commonly used retrieval measures are mean/median rank, precision, recall@k and top-n accuracy. It is a frequently used for temporal prediction tasks where the search space is finite or a variable number of possible correct answers without the explicit need of the labels (a more convenient measure than multi-label classification).   
  \paragraph{Text/Visual Generation}
  style evaluation is appropriate for downstream tasks where given a query or an initial text snippet, the task is to generate/complete the text or image frame(s) demonstrating prediction of the model for a reasoning task at the hand. BLEU, METEOR, and CIDEr are standard text generation metrics whereas inception score, fréchet inception distance, and R-precision are commonly used for image generation. Majority of temporal dependency tasks (e.g. commonsense, change caption generation) and a significant number of temporal prediction tasks (e.g. next frame prediction given a video) use generation measures.  
  \paragraph{Structured Sequence Generation} is used for goal-driven action prediction tasks such as planning and vision language navigation. Provided the initial and the goal state, the model's capability to identify a sequence of actions that can transform the initial state into the goal state is evaluated. Sequence Accuracy, Predicted/Success Path Length, Navigation Error, Task Success Rate, Goal-condition/Sub-goal Success Rate are common measures for sequence generation. 
  
%   \paragraph{Human Performance} on a dataset provides indication about quality of the dataset (specifically clarity in terms of the task and language constructs used) and considered as an upper bound of the performance one can achieve on a dataset.
   
  % \paragraph{Generalization and Robustness Experiments}

%   \mpnote{That said, there are always some limitations associated with different evaluation metrics. For example, in the case of textual and visual commonsense tasks, regular metrics can not cover all the possibilities of the system being correct. Therefore, evaluations are done on limited available knowledge. In the case of real-life, there can be any number of valid possibilities whether it is commonsense tasks or planning based tasks. Although recent advancements on simulations are able to address this with other limitations such as compute and time efficiency. Hence, the question is how can we evaluate system's creativity or commonsense of the world. To answer this question, we need better and efficient benchmark dataset and evaluation methods.}

\vspace{-0.2cm}
\section{Methods and Performance}
\label{sec:mod}
% \begin{itemize}
    
    In this section, we discuss various techniques and models that have been developed to address action-effect reasoning task. We summarize  state-of-the-art (SOTA) performance (to the best of our knowledge) on each dataset described in \ref{sec:datasets} and human upperbound (if reported) in Table \ref{table:sota}.
    
     \paragraph{Natural Language Processing Models} have  challenging benchmarks involving commonsense reasoning on the physical world around us. Some previously proposed approaches involve predicting the action-state transition and creating knowledge graphs involving physical properties, respectively for ProPara and VerbPhysics benchmarks. However, self-attentions-based methodologies such as BERT/T5/GPT are being applied across the different reasoning tasks. For example, UnifiedQA is achieving human-level performance on PIQA \cite{khashabi2020unifiedqa}.

     \paragraph{Vision+Language Models} can be treated as the core of commonsense reasoning. Here as well, transformer-inspired models are designed with a new set of pre-training strategies depending upon the tasks. \cite{sun2019videobert} proposed the VideoBERT for learning video+text embedding, which is then used for downstream tasks such as action recognition/localization. While, another transformer-based approach, named Aloe, was proposed by \cite{ding2021attention} to perform the reasoning over simulated object trajectories involving counterfactual action consequences imagination. However, when it comes to Vision Language Navigation-based tasks, we have a different set of approaches developed. \cite{blukis2022persistent} proposed the Hierarchical Language-conditioned Spatial Model (HLSM), which involves then two controllers one to generate the sub-goals and another one to generate the sequence of actions to achieve the goal. Most of the pre-training-based models involve the language encoder to handle the textual input. However, some approaches focus on learning the concept vectors which can be used to predict the outcome of a particular action on either images or videos \cite{chen2021grounding}. \cite{liu2021learning} proposed a method that focuses on instructions by modeling the actions as an energy function and predicting the next state as a composite energy function. In VisualComet \cite{park2020visualcomet}, a vision-language transformer was trained that takes a sequence of inputs (image, event, place, masked inference) which predicts masked tokens of inference in a language-model style. \cite{liu2021learning} approaches the task of describing a visual scene by understanding the relational scene description in contrast to using multi-modal language and vision language models. They propose to factorize the overall scene description for each relation to improve the overall generation and editing of scenes with multiple sets of relations. 

    % \paragraph{Vision + Language} specific benchmark involves mainly reasoning over temporal aspects, where system needs to learn the action-state and its consequences relations either from static images or from the video. Here as well, transformer inspired models are designed with new set of pre-training strategies depending upon the tasks. cite xx proposed the VideoBERT for learning video+text embedding, which in tern is used for downstream task such as action recognition/localization. Furthermore, another transformer based approach, named Aloe, is proposed by cite xx for action recognition on synthetic and much more complex environment.
    
    % \paragraph{Computer Vision Models} specific benchmark involves mainly reasoning over temporal aspects, where system needs to learn the action-state and its consequences relations either from static images or from the video. Again, Aloe is performing state-of-the-art for action recognition on synthetic and much more complex environment. %(i.e., CATER). NOT SURE WHAT TO KEEP HERE. 
    
    \paragraph{Other Models} That said, there is another set of modeling approaches involving the interaction on closed environments or synthetic environments. \cite{lohmann2020learning} proposed an approach that involves interacting with objects to figure out the relative mass of the objects. \cite{gaddy2019pre} This is achieved using two semi-supervised phases - an environment learning phase where the agent is allowed to learn the representation of the environment by sampling language-free state transition actions. The second phase is called the language learning phase where the agent is given linguistic instructions and their outcomes on the environment. \cite{chang2020procedure} proposed the use of procedure planning in instructional videos to train agents to perform complex tasks in real-world scenarios. The authors undertake the task of determining a sequence of goal-conditioned actions %that are required 
    to achieve a visual goal given a current visual observation of the world. 
    
    % \item Classical AI methods 
    % \item Neural network based approaches\\
    % \mpnote{Neural network based approaches are widely used across the fields and domains. With the advert of self-attentions, the transformer based approaches has shown its potential for action recognition/understanding and respective consequences predictions by achieving state-of-the-art performance on various benchmarks such as YouCookII, CLEVRER, CATER etc. \cite{ding2021attention,sun2019videobert}.}\\
    % \mpnote{Because of the limitations of transformer based approaches in terms on dynamic reasoning, recently neuro-symbolic approaches are gaining more and more attentions (especially, when it involves the physical reasoning) \cite{yi2019clevrer,chen2021grounding}. These methodologies mainly uses wither the graph neural networks or physics engine to derive the action-states and the future consequences given a task definition. Although these methods are performing really well, they cannot be useful on real-life environment where there are no limits on possibilities.}\\
    % \mpnote{While there are other methods which focuses on instructions by incorporating different concept learner modules or modeling the actions as an energy function \cite{chen2021grounding,liu2021learning}. Such unique methods are designed for future action state predictions either from the initial frame of reference image or for counterfactual thinking.}
%     \item Pre-trained models 
% \end{itemize}

%{\textcolor{green}{need to add more works here, specifically discuss performance and how much they are far from human upperbound}}

\section{Key Findings and Future Scope}

In this section, we summarize our observations about  existing literature that focus on reasoning about actions.   
%\paragraph{Importance of the survey from Application point of view}

\paragraph{Synthetic Benchmarks for Action Reasoning}

For action-related datasets, it makes sense to include natural images and natural videos as they are much grounded in the real-world situations that an AI agent will encounter on a regular basis. However, collecting such datasets demand costly human annotations, which is also time-consuming. Also, in real-world data, some actions are expected to occur more frequently than other ones which creates class imbalance problem. %Due to imbalance of instances across different classes, models are prone to over-classify the majority group. 
For above two concerns, many researchers have opted for simulated data creation. About one third of the datasets we surveyed incorporates synthetically generated visual scenes. In addition to being inexpensive, scalable, and minimizing the biases, synthetic datasets allow us to keep track of ground-truth object locations and attributes (i.e. visual oracle). This encourages explainable model development. %However, this approach can only have a limited set of pre-defined actions which might not be representative of the real-world. 

\paragraph{Role of Multi-modality}
Multi-modal learning aims to build models that can process and relate information from two or more modalities. Image-text multi-modality has received significant interest in AI community recently as it is an important skill for humans to perform day-to-day tasks. Perception systems can be leveraged to identify variety of visual information and a concise way to learn through observations i.e. learn to identify or perform actions. On the other hand, language provides an effective way to exchange thoughts, communicate, query or provide justifications e.g. explaining the choice of actions while performing a task or highlighting preconditions or commonsense before performing actions. Thus, multi-modal contexts play an important role in understanding actions and reasoning about them. As highlighted in \cite{sampat2020visuo},  multi-modality is important while developing AI models. The presence of multiple modalities provide natural flexibility for varied inference tasks.%simultaneously making the reasoning process more complex as information is now spanned across and requires cross-inferencing. 

\paragraph{Infinite Knowledge and Reporting Bias}
Another relevant concern associated with the commonsense knowledge is reporting bias. Most commonsense is assumed to be known rather than documented. The resulting situation is termed as reporting bias \cite{gordon2013reporting}, which prevents models to learn various commonsense notions due to its absence in text-corpora the models are trained on. There have been efforts by \cite{sap2019atomic,park2020visualcomet,speer2017conceptnet} towards systematic collection of  large-scale commonsense databases. %It is important to note that models that utilize such knowledge sources can only answer the questions for which the required knowledge is present. Practically, it is impossible to have written knowledge about each object corresponding to each attribute and action.

\paragraph{Generalization/Robustness Experiments}  Humans are capable of performing previously learned actions with respect to new set of objects or in a novel environments/scenarios. Despite the recent advances in computer vision, natural language processing (and machine learning in general), models have been data-hungry and yet fragile. Therefore, many works incorporate generalization and robustness related ablations. Most popular ones include- cross-domain evaluations (training on one domain and testing on other), multi-hop/more complex reasoning (complexity of reasoning at test time is higher than what the model is trained for), transfer learning (knowledge gained while solving one problem needs to be applied for solving different but related problem) or view variations  (pair of images are captured from different viewing angles). Among the datasets we considered in Section \ref{sec:datasets}, we observed \cite{yang2021visual,Gokhale2019CVPRWorkshops,sampat2021clevr_hyp,chen2020graph,shridhar2020alfred,gao2018action} to incorporate one or more of aforementioned generalization aspects. %Inspired by this, often times generalization and robustness related ablations are performed for models, specifically in new benchmarks. 
   
%\mpnote{Recent benchmarks and experiments tell that we have come far ahead than we were a few years ago. However, still, we are nowhere close to fully capable action or commonsense reasoning. There are some very challenging benchmarks have been proposed where human-level reasoning seems impossible, even with recent advancements. Some benchmarks are focusing on real-world data distributions based reasoning. However, they are not utilizing the full capacity of the action reasoning and lacks in terms of usefulness, mainly because of the requirement of human annotations. Because of that focus has shifted towards simulated data creation. Recent benchmarks on simulations have shown new unimaginable possibilities for action-reasoning in various applications, whether it is action recognition or action planning. Another trend that is showing its importance is multi-modalities. }\\

%(Liked this one, just needs some linguistic polishing)!

%{\textcolor{green}{what are -- very challenging benchmarks have been proposed where human-level reasoning seems impossible? gotta cite them}}

%\section{Future Scope}

\paragraph{Multi-hop Action Reasoning}
Multi-hop reasoning task refers to drawing conclusions about two or more contexts. For example, in order to assess whether `jumping into the middle of an active volcano' makes sense. Even though we have not encountered such a situation anytime, our brain figures out relevant associations from the memory- lava is hot, hot things burn, burns are painful, painful things are bad. Using this reasoning chain, we are able to conclude that it is not a good idea. While we try to develop autonomous agents with such a capability, it is important that they are able to draw such conclusions using multi-hop reasoning. 

\paragraph{Unified Action Understanding Framework}
In Section \ref{sec:mod}, we showed that existing approaches have achieved reasonably good performance on existing benchmarks. Though there exists a wide range of tasks that focus on different aspects of reasoning about actions that are independently being solved, they do not benefit each other. The best models are still dataset-specific and lack consolidated understanding of actions that are necessary to demonstrate robustness and generalization in real-world applications. As pointed out by \cite{davis2015commonsense}, the problem of integrating action descriptions at different levels of abstraction is still an open challenge. While models might understand different actions in isolation, reasoning about how they interrelate is substantially unsolved. For example, ``if you slice a finger while cutting, your parents-in-law may not be impressed". 
In this regard, we highly recommend creating an umbrella task which supports end-to-end understanding about actions- from understanding action intents and pre-conditions to imagining outcomes of hypothetical situations and demonstrating social commonsense.

\paragraph{Example:} Action ``cutting a potato" has following social commonsense (not exhaustive) associated with it;
\begin{itemize}[leftmargin=*]
    \item \textbf{Objects} \\
- Cutting requires a sharp object which is typically a knife \\
- Optionally one can have a chopping board (as a surface to cut on), plate or a bowl (to store once potatoes are cut)
 
 \item \textbf{Preconditions} \\
- Potato should be solid (in order to be cut) \\
- Knife should be sharp (in order to cut) \\
- Potato cutting normally takes place in the kitchen

 \item \textbf{Correct Effects} \\
- Potato slices/cubes/strips will be obtained depending on how they are cut

 \item \textbf{High-level Goals} \\
- Pieces of potato can be used as a component to prepare a meal and one can eat that

 \item \textbf{Undesirable Situations} \\
- While cutting a potato, one can cut him/herself  \\
- While cutting a potato, the potato has gone bad (worm inside, fungus, stinky etc.)

 \item \textbf{Undesirable Situations Effects} \\
- If one cuts their finger with a knife, it will bleed, hurt etc. \\
- If potato looks bad, it should be discarded partially or fully 
 
 \item \textbf{Desirable Post-conditions} \\
- It is recommended to soak the cut potatoes in water if there is a gap between cutting and using them

 \item \textbf{Desirable Post-conditions Effects}\\
- Soaking potatoes in water prevents them from blackening in presence of air or removes excess starch from them

 \item \textbf{Temporal Reasoning}\\
- Peel potato (optional) typically takes place before `cut potato' action which is followed by `fry/roast/bake potato'

\end{itemize}

%\paragraph{Hypothetical Reasoning about Actions}
% For example, the process of cooking dinner may involve actions such as ``Getting to know my significant other’s parents", ``Cooking dinner for four", ``Cooking pasta primavera" and ``Chopping a zucchini". 

%\paragraph{Rule-changing Action Domains}
% AI based game-playing has been an independent and well explored research direction. Given a limited set of rules, current AI methods have been able to close the gap between model performance and human-level. Over the time, researchers have been interested in incorporating language into games and developing better models that can adapt to constantly changing conditions in the games. DARPA's SAIL-ON \cite{WinNT}  

% Overall, we believe that having autonomous systems equipped with a capability to reason about actions  will further advance AI research.

\section{Conclusion}
Intelligent systems that interact with humans and assist them in performing tasks require two key abilities- understanding the environment using highly correlated vision+language understanding and performing relevant actions to achieve goals. Often times, such systems operate in open and highly uncertain environments for which having physical commonsense, understanding of goal-driven action-effects, and performing hypothetical reasoning are essential. Reasoning about actions is a relatively new research direction for the vision+language community; still there exists an array of well-defined tasks, datasets to start with. Though existing approaches can achieve reasonably good performance on carefully crafted benchmarks, they lack a consolidated understanding of actions that is necessary to demonstrate robustness and generalization to be useful in real-world applications. We discuss several directions that have potential and are unexplored, to the best of our knowledge. By presenting this survey, we aim to open up promising avenues in vision+language research and enhance relevant applications.  

\newpage
\newpage
\bibliographystyle{named}
\bibliography{ijcai22}

\end{document}